\documentclass[a4paper,fleqn]{cas-dc}
\usepackage[numbers]{natbib}

\usepackage{color,soul}
\usepackage{url}
\usepackage{algorithm}
\usepackage{algorithmic}
\usepackage{amsmath}
\usepackage{subfigure}
\usepackage{setspace}
\usepackage{amssymb}
\usepackage{pifont}
\usepackage{enumitem}

\newtheorem{Problem}{Problem Definition}

\def\tsc#1{\csdef{#1}{\textsc{\lowercase{#1}}\xspace}}
\tsc{WGM}
\tsc{QE}
\tsc{EP}
\tsc{PMS}
\tsc{BEC}
\tsc{DE}
\begin{document}
\let\WriteBookmarks\relax
\def\floatpagepagefraction{1}
\def\textpagefraction{.001}
\shorttitle{MeSIN}
\shortauthors{Yang An et~al.}

\title [mode = title]{MeSIN: Multilevel Selective and Interactive Network for Medication Recommendation}

\author[1]{Yang An}
\credit{Conceptualization, Methodology, Software, Validation, Writing-original draft, Formal Analysis, Data Curation}

\address[1]{School of Computer Science and Technology, Dalian University of Technology, Dalian, China, 116024}

\author[2]{Liang Zhang}
\credit{Supervision, Writing - Review \& Editing}
\address[2]{International Bussiness College, Dongbei University of Finance and Economics, Dalian, China, 116025}

\author[3]{Mao You}
\credit{Supervision}
\address[3]{Department of Health Technology Assessment, China National Health Development Research Center, Beijing, 100001, PR China}

\author[3]{Xueqing Tian}
\credit{Supervision}

\author[4]{Bo Jin}[orcid=0000-0002-4094-7499]
\cormark[1]
\ead{jinbo@dlut.edu.cn}
\credit{Supervision, Funding acquisition}
\address[4]{School of Innovation and Entrepreneurship, Dalian University of Technology, Dalian, China, 116024}

\author[1]{Xiaopeng Wei}
\cormark[1]
\ead{xpwei@dlut.edu.cn}
\credit{Project administration}


\begin{abstract}
Recommending medications for patients using electronic health records (EHRs) is a crucial data mining task for an intelligent healthcare system. It can assist doctors in making clinical decisions more  efficiently.
However, the inherent complexity of the EHR data renders it as a challenging task: (1) \textit{Multilevel structures}: the EHR data typically contains multilevel structures which are closely related with the decision-making pathways, e.g., laboratory results lead to disease diagnoses, and then contribute to the prescribed medications; (2) \textit{Multiple sequences interactions}: multiple sequences in EHR data are usually closely correlated with each other; (3) \textit{Abundant noise}: lots of task-unrelated features or noise information within EHR data generally result in suboptimal performance. To tackle the above challenges, we propose a multilevel selective and interactive network (MeSIN) for medication recommendation. Specifically, MeSIN is designed with three components. First, an attentional selective module (ASM) is applied to assign flexible attention scores to different medical codes embeddings by their relevance to the recommended medications in every admission. Second, we incorporate a novel interactive long-short term memory network (InLSTM) to reinforce the interactions of multilevel medical sequences in EHR data with the help of the calibrated memory-augmented cell and an enhanced input gate. Finally, we employ a global selective fusion module (GSFM) to infuse the multi-sourced information embeddings into final patient representations for medications recommendation. To validate our method, extensive experiments have been conducted on a real-world clinical dataset. The results demonstrate a consistent superiority of our framework over several baselines and testify the effectiveness of our proposed approach.
\end{abstract}


\maketitle
\section{Introduction}
\label{Introduction}

\begin{figure}
\centering
    \includegraphics[width=1\linewidth]{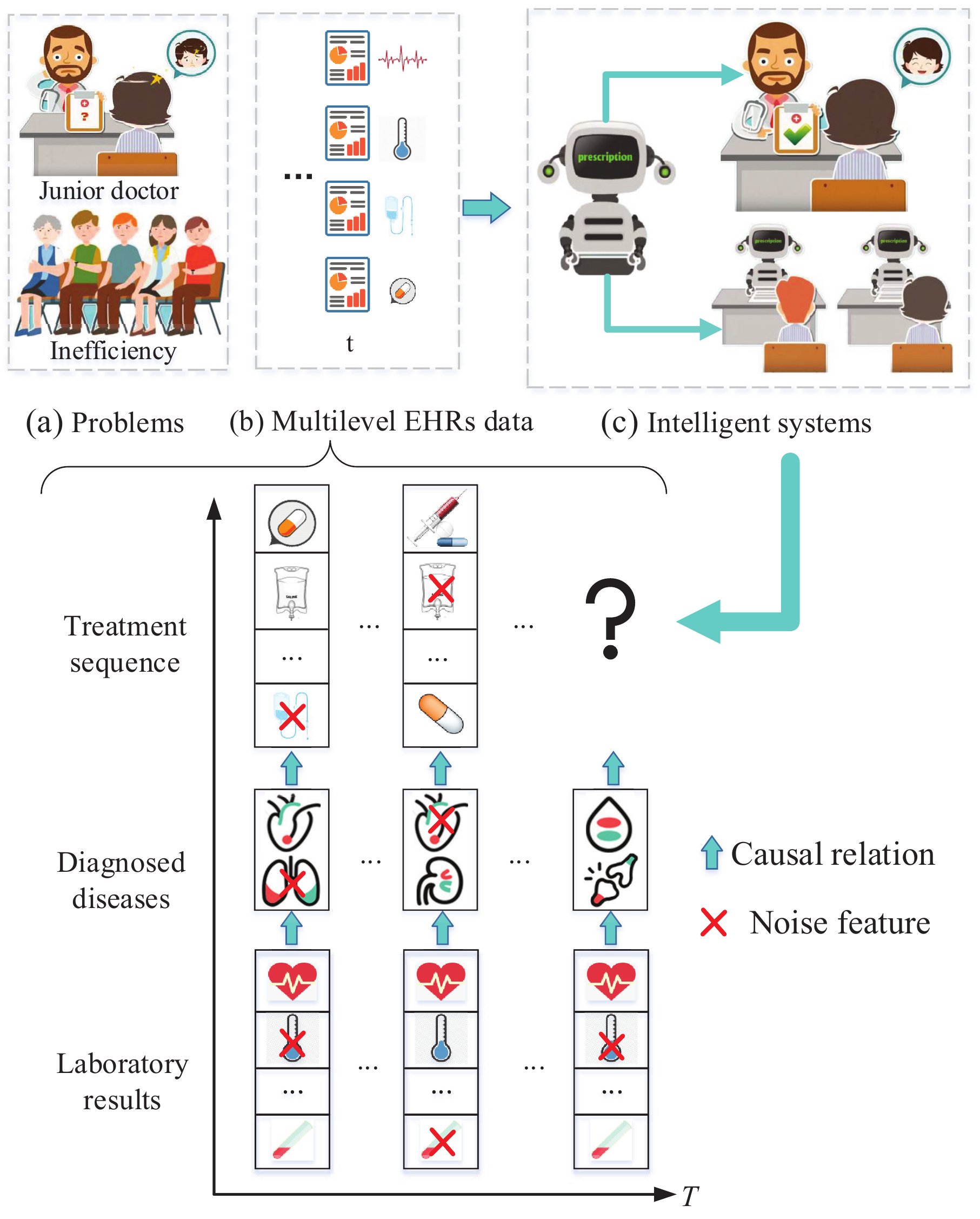}
    \caption{The urgent need of developing intelligent system with multilevel EHRs data, and corresponding complexity.}
    \vspace{-0.2cm}
    \label{Fig:Introduction}
\end{figure}

Recently, healthcare intelligence has become a hot research topic, which is mainly due to the following factors: 1) the wide utilization of digital healthcare systems that produce huge valuable data such as electronic health records (EHRs); 2) the tremendous advancements of computational models, in particular the deep learning methods; 3) an urgent need of intelligent healthcare systems to assist the junior doctors and solve the inefficiency of medical resources (Figure \ref{Fig:Introduction} (a)), brought by the emergent public health incidents, such as COVID-19 or Coronavirus Pandemic \cite{He2020TemporalDI}.
One of the core EHR-based applications is recommending medications for patients, with the aim to assist or even replace doctors in making effective and safe medication prescriptions for certain patients, as shown in Figure \ref{Fig:Introduction} (c).

However, recommending medications for patients is a challenging task due to the complexity of EHR data.
As illustrated in Figure \ref{Fig:Introduction} (b), this complexity can be attributed to several factors.
First, the EHR data typically comprises of multilevel medical records including three key aspects, e.g., laboratory results, diagnosed diseases, and prescribed treatment medications. Within each visit, the multilevel structure is closely related with the decision-making pathway, which is a kind of hierarchical structure. As shown in Figure \ref{Fig:Introduction} (b), the hierarchy generally begins with the laboratory results that precisely record the detailed health progression of a patient, the middle is the diseases diagnosed by doctors according to corresponding laboratory results, and the top is the medications prescribed by doctors after comprehensive decision-making processes.
Thus, how to fully leverage the inherent multilevel structural information has become a critical factor for modeling the intelligent medication recommendation systems.
Most existing medication recommendation studies \cite{Shang2019GAMENetGA,Zhang2017LEAPLT,He2020AttentionAM} put more efforts on modeling the mapping relations between diagnosed diseases and recommended medications. Though these algorithms have achieved early success on the medication recommendation task, they often over-emphasize the visit-level temporal dependency, and overlook the critical influence of such a hierarchy shown in Fig.\ref{Fig:Introduction}.

Second, along with the temporal dependencies of multiple medical sequences, the complex sequential correlations embodied in the multilevel structure of EHR data (Figure \ref{Fig:Introduction} (b)) is another challenge for a medication recommendation task.
For example, the laboratory results can provide enough hints for certain diseases, e.g., when the anion gap is abnormally high, creatinine is abnormally low, and aspartate is abnormally high, it demonstrates some pulmonary related diseases including respiratory infections, pulmonary emphysema, and the patient needs corresponding treatments such as glucose, sodium bicarbonate, xylitol and budesonide. Such phenomenon clearly indicates multilevel correlations of EHR data.
While most existing methods \cite{Shang2019GAMENetGA,Zhang2017LEAPLT,Le2018DualMN} overlook such important relations of medical sequences and only consider the temporal dependencies.
Though LSTM-DE \cite{Jin2018KDD} and RAHM \cite{AN2020103502} model the interactions of two sequences, they only considered the effects of related input sequences on the memory cell state while neglecting the influence on the current input state.
Thus, in this paper, we consider infusing the interactions of two sequences both on the memory cell state and input cell state simultaneously into the temporal sequence learning network.

Third, unlike the above discussed structure-related limitations, how to recognize and filter out noisy information existing in EHR data at each timestamp is another important challenge that inhibits the recommendation performance.
However, few deep learning studies in health informatics focus on infusing the feature selections into the learning process except LSAN \cite{Ye2020LSANML} which considers assigning flexible attention weights to different diagnosis codes via their relevance to corresponding diseases for reducing the effect of irrelevant diagnosis codes in EHR data.
However, the doctors practically pay more attention on the critical few factors and neglect the irrelevant medical indicators or historical medical codes.
In other words, irrelevant features should be deleted in the decision-making process, and unimportant historical medical codes should be given less attention.
In this way, the general attention mechanism might not be appropriate in the learning process.

To address the aforementioned challenges, in this paper, we develop a Multilevel Selective and Interactive Network, called MeSIN.
The key idea lies in three aspects.
First, a multilevel learning framework is designed to encode the the inherent multilevel structure of EHR data,  which imitates the decision-making process of doctors in hospitals.
Second, to capture the intra-correlations of multiple visits within each medical sequence and the inter-correlations of multiple sequences of EHR data, we propose a novel interactive temporal sequence learning network.
Third, due to the multiple heterogeneous inputs including medical codes embeddings and learned laboratory results embeddings, we introduce multiple attentional selective modules into the framework to make automatic and intelligent selections.
Therefore, our developed framework MeSIN consists of three key components including the attentional selective module (ASM), the interactive long-short term memory network (InLSTM), and the global selective fusion module (GSFM). In MeSIN, they tightly work together and significantly enhance each other for medication recommendation.

The main contributions of this study are as follows:
\begin{itemize}
    \item \textbf{Multilevel Selective and Interactive Network (MeSIN)}. To the best of our knowledge, MeSIN is the first to formulate medications recommendation task as a multilevel learning framework, which is a challenging process in clinical decision-making systems. It can fully leverage the inherent multilevel structure of EHR data to learn a comprehensive patient representation for reasonable medication recommendation.
    \item \textbf{Interactive Long-Short Term Memory Network (InLSTM)}.
    InLSTM can effectively reinforce the interactions of multiple temporal heterogeneous sequences with the help of a recurrent neural structure, a new calibrated memory-augmented cell and a novel enhanced input gate.
    \item \textbf{Attentional Selective Module (ASM)}. We incorporate multiple improved attentional selective modules into MeSIN, which can intelligently assign relevance scores to the learned medical codes embeddings according to their importance with recommended medications.
    \item \textbf{Global Selective Fusion Module (GSFM)}. We design a self-attention based global selective fusion module (GSFM) to effectively infuse the obtained heterogeneous embeddings into patient representation according to their respective importance and minimize the adverse effects induced by the irrelevant information.
\end{itemize}

\section{Related works}
\label{sec:relatedWorks}
Related studies in healthcare informatics are reviewed from the following three perspectives: medication recommendation, attention mechanism in health informatics, sequence modeling in health informatics.

\subsection{Medication recommendation}
Recently, artificial intelligence, particularly computational intelligence and machine learning methods and algorithms, has been naturally applied in the development of recommender systems to improve prediction accuracy \cite{Zhang2020ArtificialII}.
Recommending rational and effective medications in time for patients, as a paramount recommendation task in health domain, has attracted great amount of studies.
Shang et al. \cite{Shang2019GAMENetGA} categorized medication recommendation-related tasks into instance-based and longitudinal sequential recommendation methods.
Instance-based methods are based on the current disease progression of patients. For example, Zhang et al. \cite{Zhang2017LEAP} formulated the medications recommendation task as a sequential decision-making problem and leveraged a recurrent decoder to model label dependency. Wang et al. \cite{Wang2019OrderfreeMC} addressed the recommendation issues by casting the task as an order-free Markov decision process (MDP) problem.
However, they all ignored valuable historical information.
Until now, longitudinal sequential recommendation methods mainly consider the impact of historical medical records by modeling their temporal dependencies. For instance, Jin et al. \cite{Jin2018ATE} developed three different LSTMs to model heterogeneous data interactions for predicting the next-period prescriptions. Shang et al. \cite{Shang2019GAMENetGA} incorporated historical diseases and procedure codes, as well as medication records, in their model. Shang et al.\cite{shang2019pre} considered hierarchical knowledge about diagnoses and medications to enhance the code representation for medication recommendation.
An et al. \cite{AN2020103502} formulated the medication prediction task as hierarchical multi-task learning framework for improving the interpretability of predicted resutls.
However, few of them simultaneously consider all the multiple heterogeneous sequences and the correlations between them in the decision-making of medications recommendation.

\subsection{Attention mechanism in health informatics}
The attention mechanism has been proposed to automatically assign importance scores according to the information relevance. In this case, larger weights indicate that the corresponding vectors are more relevant to generating the output. Due to its powerful ability, the attention mechanism has been widely used in various neural network based applications such as language understanding tasks \cite{devlin_etal_2019_bert},\cite{Pruthi2020LearningTD}, computer vision problems \cite{Shen2019SharpAN},\cite{Hu2021AttentionalKE}.
Likewise, attention mechanism in health informatics has been prevalent in predictive modelling. For instance, GRAM \cite{Choi2017GRAMGA}, KAME \cite{Ma2018KAMEKA}, and G-BERT \cite{shang2019pre} leveraged the attention mechanism to integrate domain knowledge into disease or medication code representations for better performance.
Retain \cite{Choi2016RETAINAI}, Dipole \cite{Ma2017DipoleDP}, Timeline \cite{Bai2018InterpretableRL} and LSAN \cite{Ye2020LSANML} all introduced attention mechanism to model the disease progression by considering the dependencies among visits and provide some interpretable insights.
In addition, GCT \cite{Choi2020LearningTG} were equipped with advanced attention networks, i.e., Transformer \cite{NIPS2017_7181}, to build the correlations between medical codes from every visits based on the automatically learned attention weights.
Likely, AMANet \cite{He2020AttentionAM} utilized multiple attention networks including self-attention and inter-attention to capture the intra-view interaction and inter-view interaction. However, the attention mechanism used in above models all generated the dense attention weights without zero weights value, which means that they cannot filter out the noise information and attend focus on the critical aspects.

\subsection{Sequence modeling in health informatics}
Due to the complexity of clinical scenarios, as shown in Fig.\ref{Fig:Introduction}, EHR systems in hospitals accumulate complex temporal and heterogeneous sequences. Existing studies in health informatics have widely utilize the temporal sequential records from EHRs to solve healthcare problems such as predicting disease progression \cite{Choi2018MiMEMM}, \cite{Qiao2019MNNMA}, \cite{Zhang2019ATTAINAT}, \cite{Ye2020LSANML}, medications recommendation \cite{Zhang2017LEAP}, \cite{Shang2019GAMENetGA}, \cite{Jin2018KDD} and clinical trial recruitment \cite{Biswal2020Doctor2VecDD}, \cite{Zhang2020DeepEnrollPM}.
However, most of the studies such as T-LSTM \cite{Baytas2017PatientSV}, MNN \cite{Qiao2019MNNMA} and LSAN \cite{Ye2020LSANML} mainly focused on modelling the temporal dependencies of multiple visits of homogeneous sequence such as the history diseases sequence.
While the medication recommendation task involves multiple temporal and heterogeneous sequences, not only the temporal intra-dependencies but also the inter-correlations between the sequences should be considered when modelling the sequence learning process.
Though GAMENet \cite{Shang2019GAMENetGA} utilized two medical sequences to model the temporal dependencies for medications recommendation, it didn't consider the correlations of sequences.
DMNC \cite{Le2018DualMN} presented a two-view sequential learning model to model the complex interactions. However, the complex differentiable neural computer (DNC) blocks used in DMNC \cite{Le2018DualMN} do not explicitly model sequential interactions.
In contrast, Jin et al. \cite{Jin2018KDD} developed three heterogeneous LSTM models to model the correlations of different types of medical sequences by connecting hidden neurons, but neglect the impact on patient's current status.
MiME \cite{Choi2018MiMEMM} modelled the inherent multilevel structures of medical codes by incorporating the relationships between the diagnoses and their treatments into patient visit representations.
AMANet \cite{He2020AttentionAM} utilized multiple attention networks to capture the intra- and inter- view interactions of heterogeneous and temporal sequences, but overlook the multilevel nature of EHR data.

\section{Methods}
\subsection{Problem definitions}
\label{PD}
To facilitate the latter introduction of our computational methods and generalize the applicable dataset, we define the data from electronic health record system (EHR) using the mathematical symbols as follows.
The longitudinal EHR data contains a large number of patient records, and each patient can be represented as a sequence of multivariate observations: $\mathcal{P} = \left\{\mathcal{X}^1,\mathcal{X}^2,...,\mathcal{X}^{t_n}\right\}$ over time, where $n \in\{1,2, \ldots, N\}$, $N$ is total number of patients, and $t_n$ is the number of visits for the $n$-th patient.
Without loss of generality, we will describe the model for a patient and the subscript (n) will be dropped whenever it is unambiguous.
Each visit $\mathcal{X}^{t}$ consists of sequential laboratory indicator-wise results $\mathcal{L}^{t} = \left\{[\mathit{l}_{11},\dots,\mathit{l}_{1T}],\dots,[\mathit{l}_{q1}, \dots,\mathit{l}_{qT}] \right\}$, where $\mathit{l}_{qT}$ denotes the $q$-th indicator result at $T$-th timestamp within $t$-th visit, and categorized data including $\mathcal{C}_d^{t} \subset \mathcal{C}_d$ (a union set of diagnoses codes) and $\mathcal{C}^{t}_m \subset \mathcal{C}_m$ (a union set of medications codes).
For simplicity, we use $\mathcal{C}^t_*$ to represent the unified definition of medical codes. $\mathcal{C}_*$ denotes the medical code set and $|\mathcal{C}_*|$ denotes the size of medical code set.
$\textbf{c}^j_*$ is the $j^{th}$ medical code in $\mathcal{C}_*$.

\begin{Problem}[\textbf{Medication recommendation}]
Given the historical visit records of a patient $\left\{\mathcal{X}^{1},\mathcal{X}^{2},...,\mathcal{X}^{t-1}\right\}$, the current laboratory results $\mathcal{L}^{t}$ and the diagnosed diseases $\mathcal{C}_d^{t}$, our goal is to recommend reasonable medications by generating the multi-label output $\hat{\boldsymbol{y}}^m_{t} \in \{0,1\}^{\left|\mathcal{C}_{m}\right|}$:
\vspace{-0.3cm}
\begin{equation}
    \hat{\boldsymbol{y}}^m_{t} = f(\left\{\mathcal{X}^{1},\mathcal{X}^{2},...,\mathcal{X}^{t-1}\right\}, \mathcal{L}^{t}, \mathcal{C}_d^{t}).
\label{target}
\end{equation}
\end{Problem}

\subsection{Multilevel selective and interactive network}
\label{sec:MeSIN}

\begin{figure*}
\centering
\includegraphics[width=1\linewidth]{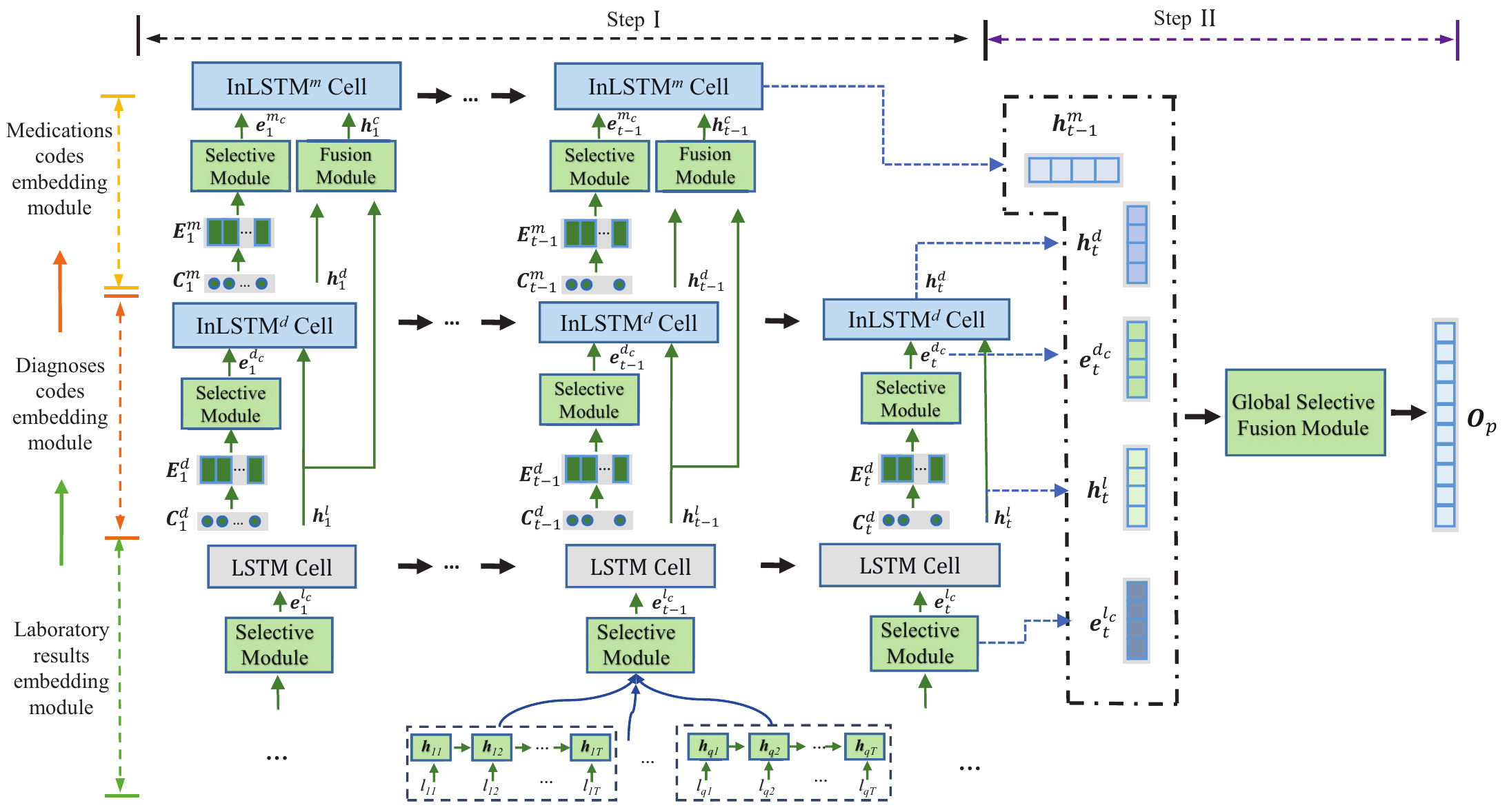}
\caption{The architecture of MeSIN. Overall, from the bottom up, MeSIN comprises of three hierarchically correlative modules and a global fusion module. The laboratory results embedding module first projects the temporal sequence of each laboratory indicator $\{\textit{l}_{q1},\dots, \textit{l}_{qT}\}$ into embedded vectors $\{\boldsymbol{h}_{1T}^l,\dots,\boldsymbol{h}_{qT}^l\}$ via a multi-channel GRU, and then uses the attentional selective module to compute the enhanced embedding $\boldsymbol{e}^{l_c}_{t}$ which would be input into LSTM to obtain the visit-level embedding $\boldsymbol{h}^l_{t}$; The diagnoses codes embedding module and medications codes embedding module respectively contain three substructures: the medical codes embedding layer for mapping the medical codes set $\boldsymbol{C}^t_*$ into dense embeddings set $\boldsymbol{E}^t_*$, the attentional selective module for computing the enhanced embedding $\boldsymbol{e}^{*_c}_{t}$, and the interactive LSTM (InLSTM) for calculating the visit-level embedding $\boldsymbol{h}^*_{t}$. Finally, the global selective fusion module is used to infuse the learned multi-sourced embeddings into patient representation $\boldsymbol{O}_p$ for medication recommendation.}
\label{Fig:Model_structure}
\end{figure*}

We propose a novel architecture, MeSIN, to implement the medications recommendation task. As shown in Fig.\ref{Fig:Model_structure}, MeSIN is a multilevel learning framework, which mainly consists of two steps. In step \textbf{\uppercase\expandafter{\romannumeral1}}, the hierarchical historical information embeddings learning process begins with the laboratory results embedding learning module, followed by the diagnoses codes embedding module, and then the medications codes embedding module. In step \textbf{\uppercase\expandafter{\romannumeral2}}, the global selective fusion module is utilized to infuse the learned heterogeneous embeddings into the patient representation according to the selective weights.

\subsubsection{Laboratory sequence embedding module}
\label{sec:lab}
As shown in Fig.\ref{Fig:Model_structure}, the module mainly consists of three key parts: a multi-channel time-series embedding layer, an attentional selective layer, and  a temporal sequence learning network.

\paragraph{\textbf{Multi-channel time-series embedding layer}.}
As the meanings of particular clinical features for patients in diverse medical conditions are different, the progression of laboratory indicators are distinct accordingly. Thus, the embedding of sequential feature representing the indicator changing progress is distinct from each other. Here, inspired by ConCare \cite{ConCare2020}, we employ the multi-channel time-series embedding layer to embed the sequence of each laboratory indicator feature separately by multi-channel GRUs:
\begin{equation}
\label{Eq:Multi_LSTM}
\begin{split}
\boldsymbol{h}_{qT}^l=\operatorname{GRU}_q\left(\textit{l}_{q1}, \textit{l}_{q2}, \dots, \textit{l}_{qT}\right),
\end{split}
\end{equation}
where $\{\textit{l}_{q1}, \textit{l}_{q2}, \dots, \textit{l}_{qT}\}$ denotes the time series and $\boldsymbol{h}_{qT}^l$ represents the embedded vectors of feature q. Therefore, all the embedded vectors of time series of indicator features $\{\boldsymbol{h}_{1T}^l,\dots,\boldsymbol{h}_{qT}^l\}$ can be acquired in the same way. To reduce clutter, the superscript (t) representing the results generated at $t$-th visit will be dropped whenever it is unambiguous.

\paragraph{\textbf{ASM for laboratory results embeddings selection}.}
For each sequence of laboratory results, we gain the corresponding embeddings ${\boldsymbol{h}_{1T}^l,\dots,\boldsymbol{h}_{qT}^l}$. However, in clinical scenarios, doctors pay attention to only a few paramount indicators according to clinical experience, which can effectively improve work efficiency.

In this case, the attention mechanism that computes the attention weights using \textbf{softmax function} \cite{Softmax} might be inappropriate, because it results in dense attention alignments that is wasteful and making models less interpretive.
Therefore, we introduce a sparse attention using \textbf{entmax} \cite{Peters2019SparseSM} computing attention weights for ASM of MeSIN to increase focus on relevant source medical codes embeddings and make the model more interpretable.
Here, we employ the specially proposed ASM to compute the enhanced laboratory results embedding $\boldsymbol{e}^{l_c}_{t}$:
\begin{equation}
\label{Eq:Lab_ASM}
\begin{split}
\boldsymbol{e}^{l_c}_{t} &= \sum_{i=1}^{q} \alpha_{i}^l\boldsymbol{h}_{iT}^l (\alpha_{i}^l\in\boldsymbol{\alpha^l}),\\
\boldsymbol{\alpha^l} &= \operatorname{\alpha-entmax}([\alpha_{1}^l,\alpha_{2}^l,\dots,\alpha_{q}^l],{\gamma}_l),\\
\alpha_{i}^l &= \operatorname{tanh}({\boldsymbol{W}_{l_a}}^\intercal\boldsymbol{h}_{iT}^l + \boldsymbol{b}_{l_a}),
\end{split}
\end{equation}
where $\boldsymbol{W}_{l_a}\in\mathbb{R}^d$ and $\boldsymbol{b}_{l_a}\in\mathbb{R}$ are the parameters of ASM to be learned. $\alpha-entmax$ \cite{Peters2019SparseSM} is a special method of entmax, by which we can find the optimal equilibrium point by controlling the value of ${\gamma}_l$. For ${\gamma}_l > 1$, as the value increases, entmax tends to produce sparse probability distributions, yielding a function family  interpolating between softmax and sparsemax.
In this way, we can compute all the enhanced laboratory results embeddings $\{\boldsymbol{e}^{l_c}_{1},\dots,\boldsymbol{e}^{l_c}_{t}\}$ at each timestamp.

\paragraph{\textbf{Temporal sequence learning network}.}
To further capture the temporal dependency of multi-visit laboratory results, the enhanced laboratory results embeddings $\{\boldsymbol{e}^{l_c}_{1},\dots,\boldsymbol{e}^{l_c}_{t}\}$ will be input into the temporal sequence learning network for combining with the historical laboratory results:
\begin{equation}
\label{Eq:Lab_visit}
\begin{split}
\boldsymbol{h}^l_{t}=\operatorname{LSTM}_L\left(\boldsymbol{h}^l_{t-1},\boldsymbol{e}^{l_c}_{t}\right),
\end{split}
\end{equation}
where $\operatorname{LSTM}_L$ represents the long-short temporal neural network (LSTM) for capturing the temporal dependency of laboratory examination sequence, $\boldsymbol{h}^l_{t}$ denotes the obtained visit-level laboratory results embedding containing the history information at $t$-th visit.
With the same calculations in the remaining timestamps, we can finally have all the history laboratory results embeddings $\{\boldsymbol{h}^l_{1},\dots,\boldsymbol{h}^l_{t}\}$ which will be input into the following embedding modules.

\subsubsection{Diagnoses codes embedding module}
\label{sec:diag}
After checking the laboratory results, the doctors tend to retrieve the history diagnosed diseases and combines them with current disease condition for comprehensive decision-making.
Likely, as shown in Fig.\ref{Fig:Model_structure}, we design a module that contains three critical parts:
a diagnosis code embedding layer, an attentional selective module and a novel temporal sequence interactive learning network.

\paragraph{\textbf{Diagnosis code embedding layer}.}
Taking the timestamp $\textit{t}$ as an example, MeSIN first encodes each diagnosis code $\boldsymbol{c}^d_i$ into a dense representation vector $\boldsymbol{e}^d_i \in \mathbb{R}^d$ as:
\begin{equation}
  \label{Eq:embedding}
  \boldsymbol{e}^d_i = \boldsymbol{W}^d_{e}\boldsymbol{c}^d_i,
\end{equation}
where $\boldsymbol{W}^d_{e} \in\mathbb{R}^{d \times |\mathcal{C}_*|}$ is the embedding matrix of medical codes that needs to be learned, $d$ is the size of embedding dimension, and $|\mathcal{C}_*|$ is the size of medical code set. Thus, for the diagnosis code set $\mathcal{C}_d$, we can represent it by a collection of dense representation vectors $\boldsymbol{E}^d = [\boldsymbol{e}^d_1,\dots,\boldsymbol{e}^d_{|\mathcal{C}_d|}] \in \mathbb{R}^{d\times|\mathcal{C}_d|}$. Then, for the $i$-th visit, we can obtain dense embedding set $\boldsymbol{E}^d_t = [\boldsymbol{e}^d_1,\dots,\boldsymbol{e}^d_m]\in \mathbb{R}^{d\times|m}$, in which each embedding vector is extracted from $\boldsymbol{E}^d$ if it exists in $i$-th visit.

\paragraph{\textbf{ASM for diagnoses codes embeddings selection}.}
However, as discussed before, not every historical disease has impact on the future disease risk, we should assign different relevance scores on each code embedding according to their importance degree. Here we also leverage ASM  for diagnoses codes embeddings selection. The enhanced diagnosis code embedding $\boldsymbol{e}^{d_c}_{t}$ can be calculated as:
\begin{equation}
\label{Eq:diag_ASM}
\begin{split}
\boldsymbol{e}^{d_c}_{t} &= \sum_{i=1}^{m} \alpha_{i}^d\boldsymbol{e}^d_i (\alpha_{i}^d\in\boldsymbol{\alpha^d}),\\
\boldsymbol{\alpha^d} &= \operatorname{\alpha-entmax}([\alpha_{1}^d,\alpha_{2}^d,\dots,\alpha_{m}^d],{\gamma}_d),\\
\alpha_{i}^d &= \operatorname{tanh}({\boldsymbol{W}_{d_a}}^\intercal\boldsymbol{e}^d_i + \boldsymbol{b}_{d_a}),
\end{split}
\end{equation}
where $\boldsymbol{W}_{d_a}\in\mathbb{R}^d$ and $\boldsymbol{b}_{d_a}\in\mathbb{R}$ are the parameters of ASM to be learned. ${\gamma}_d$ is the hyper-parameter of $\alpha-entmax$ \cite{Peters2019SparseSM} in this module.
In this way, we can compute all the enhanced diagnoses codes embeddings $\boldsymbol{E}^{d_c} = \{\boldsymbol{e}^{d_c}_{1},\dots,\boldsymbol{e}^{d_c}_{t}\}$.

\paragraph{\textbf{InLSTM in diagnosis code sequence learning}.}
In addition to modeling of the temporal dependency of a single sequence, we should also consider the interactions of multiple sequences in the sequence learning network.
As before, the laboratory results could be regarded as critical references when making the diagnoses by doctors. Hence, the sequence of laboratory results should be used to control the diagnosed disease sequence learning process.
Therefore, such kind of network will adopt two input sequences: one is the primary input $\boldsymbol{x}^{p}_{t}$ of sequence learning network such as the gained diagnosis code embedding $\boldsymbol{e}^{d_c}_{t}$, another is
the auxiliary input $\boldsymbol{x}^{a}_{t}$ for assisting in controlling the primary sequence learning process such as the learned visit-level laboratory results embedding $\boldsymbol{h}^l_{t}$.

\begin{figure}
\centering
\includegraphics[width=1\linewidth]{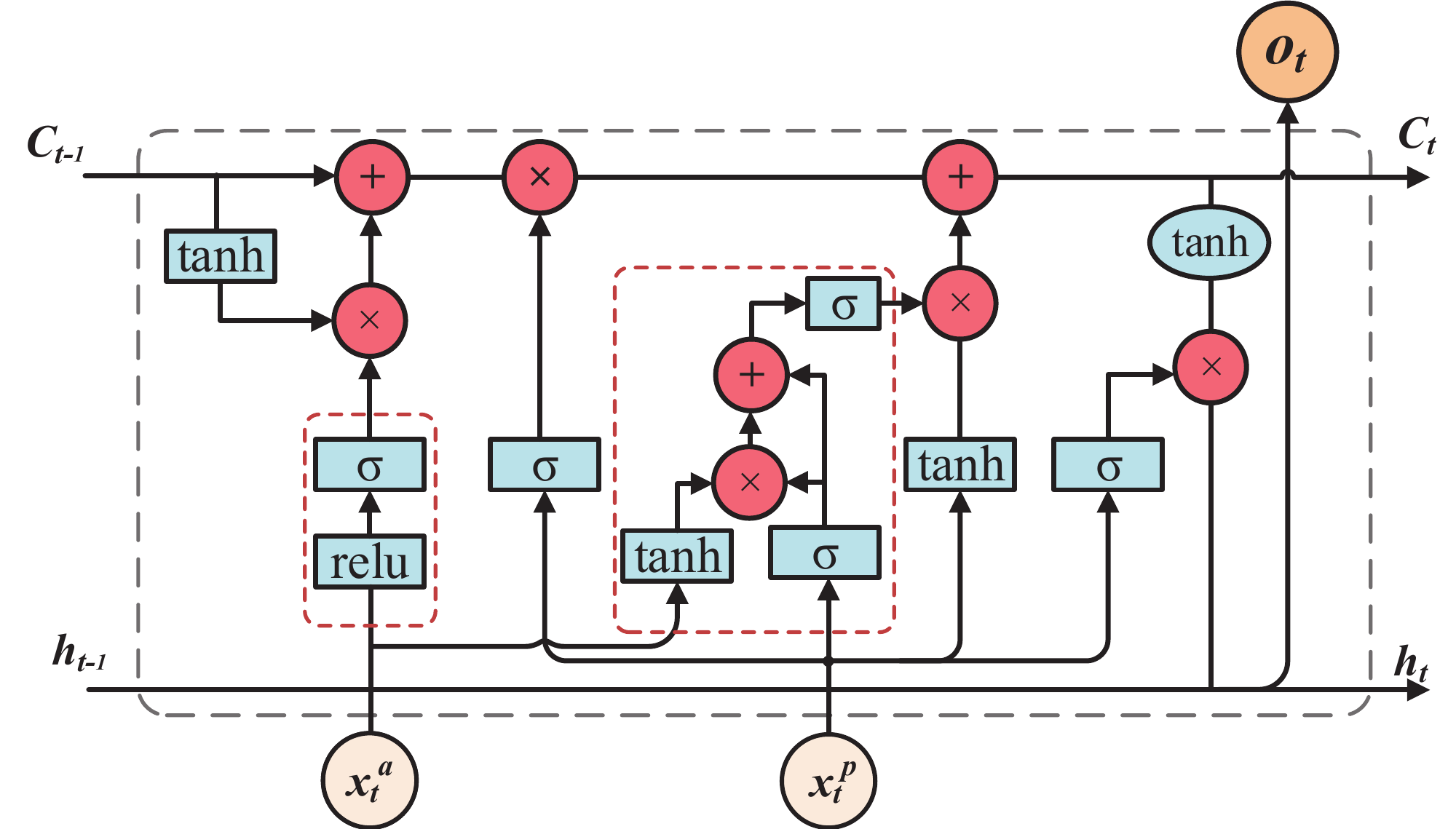}
\caption{The structure of InLSTM.}
\vspace{-0.2cm}
\label{Fig:InLSTM}
\end{figure}

Therefore, the basic LSTM model \cite{hochreiter1997long} is not appropriate under such circumstances. Inspired by LSTM-DE \cite{Jin2018ATE}, we propose a novel interactive long-short term memory network (\textbf{InLSTM}), as shown in Fig.\ref{Fig:InLSTM}, to reinforce the interaction process of two associated sequences, which brings in two novel components including a calibrated memory-augmented cell and an enhanced input gate. It can be defined as:
\begin{equation}
\label{Eq:InLSTM}
\begin{split}
\textbf{h}_{t}=\operatorname{InLSTM}\left(\textbf{h}_{t-1}, \textbf{x}^p_t, \textbf{x}_t^a\right),
\end{split}
\end{equation}
where the detailed mathematical expression of $\operatorname{InLSTM}$ is:
\begin{equation}
\label{Eq:InLSTM_Gate}
\begin{split}
\left[\begin{array}{c}\tilde{\boldsymbol{C}}_t \\ \boldsymbol{o}_{t} \\ \boldsymbol{i}_{t} \\\boldsymbol{f}_{t}\end{array}\right]
&=\left[\begin{array}{c}\tanh \\ \sigma \\ \sigma \\ \sigma\end{array}\right]
\left(
\boldsymbol{W}
\left[\begin{array}{c}\boldsymbol{x}_t \\ \boldsymbol{h}_{t-1}\end{array}\right]
+
\boldsymbol{b}
\right),\\
\textbf{C}_t &= \boldsymbol{f}_t*(\boldsymbol{C}_{t-1} +\tilde{\boldsymbol{C}}^e_t)+\boldsymbol{i}^e_t*\tilde{\boldsymbol{C}}_t,\\
\boldsymbol{h}_t &=\boldsymbol{o}_t*\tanh(\boldsymbol{C}_t),
\end{split}
\end{equation}
where $\tilde{\boldsymbol{C}}^e_t$ denotes the calibrated memory-augmented cell state calculated through Eq. (\ref{Eq:InLSTM_memory}), and $\boldsymbol{i}^e_t$ represents the enhanced input gate which can be computed through Eq. (\ref{Eq:InLSTM_input}):
\begin{equation}
\label{Eq:InLSTM_memory}
\begin{split}
\boldsymbol{d}_t &=\tanh(\boldsymbol{W}_{enh}\boldsymbol{C}_{t-1} + \boldsymbol{b}_{enh}), \\
\boldsymbol{d}^e_t &= \sigma(\boldsymbol{U}_{d}^r\operatorname{ReLU}(\boldsymbol{W}_d^r\boldsymbol{x}_{t}^{a}), \\
\tilde{\textbf{C}}^e_t &= \boldsymbol{d}_t * \boldsymbol{d}^e_t, \\
\end{split}
\end{equation}
\vspace{-0.2cm}
\begin{equation}
\label{Eq:InLSTM_input}
\begin{split}
\hat{\boldsymbol{i}}^e_t &= \operatorname{tanh}(\boldsymbol{W}_i^e\boldsymbol{x}_{t}^{a}+\boldsymbol{b}_i^e), \\
\tilde{\boldsymbol{i}}^e_t &= \boldsymbol{i}_{t} * \hat{\boldsymbol{i}}^e_t,\\
\boldsymbol{i}^e_t &= \sigma(\tilde{\boldsymbol{i}}^e_t+\boldsymbol{i}_{t})\\
&=\sigma((\boldsymbol{i}^e_t+1)\boldsymbol{i}_{t}),\\
\end{split}
\end{equation}
where $\boldsymbol{d}_t$ indicates the obtained history memory state, $\boldsymbol{d}^e_t$ denotes the calibrated gate calculated with auxiliary input $\boldsymbol{x}_{t}^{a}$.
Afterwards, $\boldsymbol{d}^e_t$ will be used to obtain the calibrated memory-augmented cell state value $\tilde{\textbf{C}}^e_t$ by multiplying the $\boldsymbol{d}_t$. By this way, the calibrated gate can selectively assign more weights to the representative and predictive memory neurons while suppressing the unimportant neurons.
For the input cell, besides the primary input $\boldsymbol{x}_{t}^{p}$ itself, it is also influenced by the auxiliary input $\boldsymbol{x}_{t}^{a}$. Under such a circumstance, the calibrated auxiliary input $\hat{\boldsymbol{i}}^e_t$ is introduced to calculate the auxiliary influence score $\tilde{\boldsymbol{i}}^e_t$ by multiplying the normal input gate $\boldsymbol{i}_{t}$. Finally, the enhanced input gate $\boldsymbol{i}^e_t$ is computed by the addition of the auxiliary influence score $\tilde{\boldsymbol{i}}^e_t$ and the normal input gate $\boldsymbol{i}_{t}$, and then adjusted to the value between 0 and 1 via a $\operatorname{sigmoid}$ function.

Therefore, for the diagnoses codes embedding module, we can calculate the final visit-level diagnoses codes embedding $\boldsymbol{h}^d_{t}$ by fusing with the historical diagnosed disease as:
\begin{equation}
\label{Eq:diag_LSTM}
\begin{split}
\boldsymbol{h}^d_{t} = \operatorname{InLSTM}_d\left(\boldsymbol{h}^d_{t-1},\boldsymbol{e}^{d_c}_{t},\boldsymbol{h}^{l}_{t}\right),
\end{split}
\end{equation}
where $\operatorname{InLSTM}_d$ denotes the proposed InLSTM (Eq.(\ref{Eq:InLSTM})) in this module, which is used to capture the correlations between the primary input of diagnosis code embedding sequence and the auxiliary input of laboratory results embedding sequence.

\subsubsection{Medications codes embedding module}
Similar to the previous hierarchy, diagnoses codes embedding module, the medications codes embedding module still comprise three main parts: a code embedding layer, an attentional selective module, and the temporal sequence interactive learning network.

\paragraph{\textbf{Medication code embedding layer}.}
In this module, MeSIN still first encodes each medication code $\boldsymbol{c}^m_z$ into a dense embedding vector $\boldsymbol{e}^m_z \in \mathbb{R}^d$ as:
\begin{equation}
  \label{Eq:med_embedding}
  \boldsymbol{e}^m_z = \boldsymbol{W}^m_{e}\boldsymbol{c}^m_z,
\end{equation}
where $\boldsymbol{W}^m_{e} \in\mathbb{R}^{d \times |\mathcal{C}_*|}$ is the embedding matrix of medical codes that needs to be learned.
Then we can gain the dense medication code embedding set $\boldsymbol{E}^m_z = [\boldsymbol{e}^m_1,\dots,\boldsymbol{e}^m_n] \in \mathbb{R}^{d\times|n}$, in which each embedding is extracted from $\boldsymbol{E}^m = [\boldsymbol{e}^m_1,\dots,\boldsymbol{e}^m_{|\mathcal{C}_m|}] \in \mathbb{R}^{d\times|\mathcal{C}_m|}$ if it existed in $i$-th visit.

\paragraph{\textbf{ASM for medications codes embeddings selection}.}
As mentioned before, MeSIN needs to filter out the noise coming from irrelevant historical medication codes sets medications at each timestamp.
In this case, we should assign different relevance scores on different codes embeddings using the attentional selective module for computing the enhanced medications set embedding $\boldsymbol{e}^{m_c}_{t-1}$:
\vspace{-0.2cm}
\begin{equation}
\label{Eq:med_ASM}
\begin{split}
\boldsymbol{e}^{m_c}_{t-1} &= \sum_{i=1}^{n} \alpha_{i}^m\boldsymbol{e}^m_i (\alpha_{i}^d\in\boldsymbol{\alpha^d}),\\
\boldsymbol{\alpha^m} &= \operatorname{\alpha-entmax}([\alpha_{1}^m,\alpha_{2}^m,\dots,\alpha_{n}^m],{\gamma}_m),\\
\alpha_{i}^m &= \operatorname{tanh}({\boldsymbol{W}_{m_a}}^\intercal\boldsymbol{e}^m_i + \boldsymbol{b}_{m_a}),
\end{split}
\end{equation}
where $\boldsymbol{W}_{m_a}\in\mathbb{R}^d$ and $\boldsymbol{b}_{m_a}\in\mathbb{R}$ are the parameters of ASM to be learned. ${\gamma}_m$ is the hyper-parameter of $\alpha-entmax$ \cite{Peters2019SparseSM} in this module.
Likely, we can compute the enhanced medications codes embeddings sequence $\boldsymbol{E}^{m_c} = \{\boldsymbol{e}^{m_c}_{1},\dots,\boldsymbol{e}^{m_c}_{t-1}\}$ at historical timestamps.

\paragraph{\textbf{InLSTM in medications codes sequence learning}.}
Finally, for capturing the temporal dependency of historical medications, the gained enhanced medication code embedding sequence $\boldsymbol{E}^{m_c}$ is treated as the primary input of sequence learning network.
In addition, recommending medications is essentially a  comprehensive decision-making process, the historical prescribed medications must be affected by the laboratory results and diagnosed diseases.
Therefore, the sequences of laboratory results and diagnosed diseases are taken as the auxiliary input assisting in controlling the sequence learning process.
Then, we can calculate the final visit-level medication code embedding $\boldsymbol{h}^m_{t-1}$ by fusing the historical disease progression using InLSTM Eq.(\ref{Eq:InLSTM}) as:
\begin{equation}
\label{Eq:med_LSTM}
\begin{split}
\boldsymbol{h}^m_{t-1} = \operatorname{InLSTM}_m\left(\boldsymbol{h}^m_{t-2},\boldsymbol{e}^{m_c}_{t-1},\boldsymbol{h}^{c}_{t-1}\right),
\end{split}
\end{equation}
where $\operatorname{InLSTM}_m$ denotes the proposed InLSTM in this module, which is used to capture the correlations between the primary input of diagnosis code embedding and the auxiliary input $\boldsymbol{h}^{c}_{t-1}$. Further, the auxiliary input $\boldsymbol{h}^{c}_{t-1}$ is calculated via a fusion module:
\begin{equation}
\label{Eq:fusion_vector}
\begin{split}
\boldsymbol{h}^{c}_{t-1} = \sigma\left(\boldsymbol{W}^d_c\boldsymbol{h}^d_{t-1}+\boldsymbol{W}^l_c\boldsymbol{h}^l_{t-1}\right),
\end{split}
\end{equation}
where $\boldsymbol{W}^d_c,\boldsymbol{W}^l_c \in\mathbb{R}^{{d}\times{d}}$, $\sigma$ denotes the activation function $\textit{tanh}$.
$\boldsymbol{h}^l_{t-1}$ and $\boldsymbol{h}^d_{t-1}$ respectively represent the obtained visit-level laboratory results embedding using Eq. (\ref{Eq:Lab_visit}) and diagnosed diseases embedding using Eq. (\ref{Eq:diag_LSTM}).

\subsubsection{Global selective fusion module}
\label{sec:GSF}
In step \textbf{\uppercase\expandafter{\romannumeral1}}, by modeling the hierarchically interactive temporal sequence learning process, we can obtain the visit-level laboratory results embedding $\boldsymbol{h}^l_{t}$, diagnoses codes embedding $\boldsymbol{h}^d_{t}$ and the medications codes embedding $\boldsymbol{h}^m_{t-1}$. All above three kinds of embeddings have incorporated corresponding historical information. For recommending medications at current timestamp, the current enhanced laboratory results embedding $\boldsymbol{e}^{l_c}_{t}$ and diagnosis code embedding $\boldsymbol{e}^{d_c}_{t}$ should be given more attention when making final decisions.

To effectively fuse above five heterogeneous embeddings according to their importance scores and minimize the effect introduced by irrelevant information as much as possible, in step
\textbf{\uppercase\expandafter{\romannumeral2}}, we design a global selective fusion module, which is realized by a self-attention mechanism.
Since the five types of embeddings are heterogeneous, here, we first calculate the information importance scores by themselves as:
\begin{eqnarray}
\label{Eq:global_att}
\begin{aligned}
  \left[
  \begin{array}{c}
  {\alpha}_{m} \\
  {\alpha}_{d} \\
  {\alpha}_{l} \\
  {\alpha}_{dc} \\
  {\alpha}_{lc}
  \end{array}
  \right]
  &=
  \left[
  \begin{array}{c}
  \boldsymbol{W}_m^{a} \\
  \boldsymbol{W}_d^{a} \\
  \boldsymbol{W}_l^{a}\\
    \boldsymbol{W}_{dc}^{a} \\
  \boldsymbol{W}_{lc}^{a}
  \end{array}
  \right]
  \ast
  \left[
  \begin{array}{l}
    \boldsymbol{h}^{m}_{t-1} \\
    \boldsymbol{h}^{d}_t \\
    \boldsymbol{h}^{l}_t \\
    \boldsymbol{e}^{d_c}_t \\
    \boldsymbol{e}^{l_c}_t
  \end{array}
  \right]
  +
  \left[
  \begin{array}{c}
  \boldsymbol{b}_m^a \\
  \boldsymbol{b}_{d}^a \\
  \boldsymbol{b}_{l}^a \\
  \boldsymbol{b}_{dc}^a \\
  \boldsymbol{b}_{lc}^a
  \end{array}
  \right],
\end{aligned}
\end{eqnarray}
where $\boldsymbol{W}_m^a,\boldsymbol{W}_{d}^a,\boldsymbol{W}_{l}^a,\boldsymbol{W}_{dc}^a,\boldsymbol{W}_{lc}^a$ and $\boldsymbol{b}_m^a,\boldsymbol{b}_{d}^a,\boldsymbol{b}_{l}^a,\boldsymbol{b}_{dc}^a,\boldsymbol{b}_{lc}^a$ are the parameters to be learned.
${\alpha}_{m},{\alpha}_{d},{\alpha}_{l},{\alpha}_{dc},{\alpha}_{lc}$ are the information importance scores, by which we can calculate the final information importance scores $\boldsymbol{a}$ as:
\begin{equation}
  \label{Eq:a_weights}
    \boldsymbol{\alpha} =  \operatorname{softmax}([{\alpha}_{m},{\alpha}_{d},{\alpha}_{l},{\alpha}_{dc},{\alpha}_{lc}]),
\end{equation}
where $\boldsymbol{a} = [{\alpha}_m^{\prime},{\alpha}_{d}^{\prime},{\alpha}_{l}^{\prime},{\alpha}_{dc}^{\prime},{\alpha}_{lc}^{\prime}]$.

Finally, we obtain the ultimate patient representation vector by summing up the heterogeneous information vectors according to importance scores from Eq.(\ref{Eq:a_weights}) as:
\begin{equation}
\label{Eq:pat_fusion}
\boldsymbol{O}_p = {\alpha}_m^{\prime}\boldsymbol{h}^{m}_{t-1} + {\alpha}_d^{\prime}\boldsymbol{h}^{d}_t + {\alpha}_l^{\prime}\boldsymbol{h}^{l}_t + {\alpha}_{dc}^{\prime}\boldsymbol{e}^{d_c}_t + {\alpha}_{lc}^{\prime}\boldsymbol{e}^{l_c}_t,
\end{equation}
where $\boldsymbol{O}_p$ is the patient representation vector to be used to recommend reasonable medications in the next subsection.

\subsubsection{Medication recommendation}
\label{sec:MR}
Doctors make decisions about recommending reasonable medications for patients after comprehensive consideration. Likewise, the learned patient representation $\boldsymbol{O}_p$ is employed in this study to recommend reasonable medications as:
\begin{equation}
\label{Eq:output}
\hat{\textbf{y}}_{t}=\operatorname{softmax}\left(\textbf{W}_{o}\cdot\textbf{O}_{p}+\textbf{b}_{o}\right),
\end{equation}
where $\hat{\textbf{y}}_{t}$ denotes the set of recommended multi-label medications,
$\textbf{W}_{o}\in\mathbb{R}^{{d_m}\times{r}}$ and $\textbf{b}_{o}\in\mathbb{R}^{d_m}$ are parameters to be learned.

\subsection{Model training}
\label{sec:model_train}
Since medication recommendation task belongs to the domain of sequential multi-label prediction task, we utilize the binary cross-entropy loss $\mathcal{L}_{ce}$ and multi-label margin loss $\mathcal{L}_{mg}$ as the objective functions.
The prediction objective function binary cross-entropy loss $\mathcal{L}_{ce}$ is formulated as:
\begin{equation}
\label{loss1}
\resizebox{.89\linewidth}{!}{$
    \displaystyle
\textcolor{black}{\mathcal{L}_{ce}=-\sum_{t=1}^{T} \boldsymbol{y}_{t} \log \sigma\left(\hat{\boldsymbol{y}}_{t}\right)+\left(1-\boldsymbol{y}_{t}\right)\log\left(1-\sigma\left(\hat{\boldsymbol{y}}_{t}\right)\right)}
$},
\end{equation}%

The corresponding objective function multi-label margin loss $\mathcal{L}_{mg}$ is:
\begin{equation}
\label{loss2}
\resizebox{.89\linewidth}{!}{$
    \displaystyle
   \textcolor{black}{\mathcal{L}_{mg}=\sum_{t}^{T} \sum_{i}^{\left|\mathcal{C}\right|} \sum_{j}^{\left|Y^t\right|} \frac{\max \left(0,1-\left(\hat{y}_{t}[Y^t_j]-\hat{y}_{t}[i]\right)\right)}{L}}
$},
\end{equation}%
where $\hat{y}_{t}[i]$ is the value of $i^{th}$ coordinate at $t^{th}$ visit and \textcolor{black}{$\hat{y}_{t}[Y^t_j]$ denotes the predicted label value indexed by $Y^t_j$, the $j^{th}$ value in the ground truth label set $Y^t$ at $t$-th visit for a patient.} For function (\ref{loss1}) and (\ref{loss2}). So we can get two binary cross entropy loss functions $\mathcal{L}^{d}_{ce}$, $\mathcal{L}^{m}_{ce}$, and two multi-label margin loss functions $\mathcal{L}^{d}_{mg}$, $\mathcal{L}^{m}_{mg}$.

To facilitate the joint optimization process of two tasks, we combine the aforementioned loss functions to build a joint loss function $\mathcal{L}$:
\begin{equation}
\label{loss_total}
\begin{split}
\mathcal{L} = \eta\mathcal{L}_{ce} + \varepsilon \mathcal{L}_{mg},
\end{split}
\end{equation}
where $\eta, \varepsilon \geqslant 0 $ are the mixture weights, and $\eta + \varepsilon = 1$.
The training algorithm is detailed in Algorithm \ref{Algorithm}.

\begin{algorithm}
\caption{Model training for MeSIN.}
\label{Algorithm}
\begin{algorithmic}[1]
\REQUIRE Training set $\boldsymbol{R}$, training epochs $N$, batch size $BS$, mixture weights $\eta, \varepsilon $ in Eq. (\ref{loss_total});\\
Use uniform distribution to initialize the model parameters $\theta \sim U(-1,1)$;
\FOR{$i = 1$ to $N * |\boldsymbol{R}|$}
     \STATE Sample $BS$ patients ($\boldsymbol{P} = \left\{\mathcal{X}^1,\dots,\mathcal{X}^{T_n}\right\}$) from $\boldsymbol{R}$;
    \FOR{$t = 1$ to ${T_i}$}
        \STATE/**Laboratory results embedding module**/
        \STATE Obtain multi-channel time-series embeddings $\{\boldsymbol{h}_{1T}^l,...,\boldsymbol{h}_{qT}^l\}$ using Eq. (\ref{Eq:Multi_LSTM});
        \STATE Obtain enhanced laboratory results embedding $\boldsymbol{e}^{l_c}_{t}$ using ASM using Eq. (\ref{Eq:Lab_ASM});
        \STATE Compute the visit-level laboratory results embedding $\boldsymbol{h}^l_{t}$ using Eq. (\ref{Eq:Lab_visit});
        \STATE/**Diagnoses codes embedding module**/
        \STATE Obtain diagnoses codes embeddings $\boldsymbol{E}^d_t = [\boldsymbol{e}^d_1,\dots,\boldsymbol{e}^d_m]\in \mathbb{R}^{d\times|m}$ using Eq. (\ref{Eq:embedding});
        \STATE Obtain the enhanced diagnosis code embedding $\boldsymbol{e}^{d_c}_{t}$ using ASM as Eq. (\ref{Eq:diag_ASM});
        \STATE Compute the visit-level diagnoses codes embedding $\boldsymbol{h}^d_{t}$ using InLSTM (Eq. (\ref{Eq:InLSTM}-\ref{Eq:diag_LSTM}));
        \STATE/**Medications codes embedding module**/
        \STATE Obtain the medications codes embeddings $\boldsymbol{E}^m_z = [\boldsymbol{e}^m_1,\dots,\boldsymbol{e}^m_n] \in \mathbb{R}^{d\times|n}$ using Eq. (\ref{Eq:med_embedding});
        \STATE Obtain the enhanced medications codes embedding $\boldsymbol{e}^{m_c}_{t-1}$ using ASM as Eq. (\ref{Eq:med_ASM});
        \STATE Compute the visit-level medications codes embedding $\boldsymbol{h}^m_{t-1}$ using InLSTM (Eq. (\ref{Eq:InLSTM}-\ref{Eq:diag_LSTM}));
        \STATE/**Global selective fusion module**/
        \STATE Incorporate the multi-source embeddings into patient representation $\boldsymbol{O}_p$ using Eq. (\ref{Eq:global_att}-\ref{Eq:pat_fusion});
        \STATE Compute recommended medications $\hat{\boldsymbol{y}}^{t}$ in Eq.(\ref{Eq:output});
    \ENDFOR
    \STATE Update $\theta$ by optimizing the total loss $\mathcal{L}$ in Eq. \ref{loss_total};
\ENDFOR
\end{algorithmic}
\end{algorithm}

\section{Experiments and discussion}
\label{EX}

\subsection{Datasets description}
\label{data}
As is analyzed in Section \ref{Introduction}, the aim of study is to recommend medications for patients based on the heterogeneous multilevel EHR data.
Hence, we should conduct experiments on a cohort where patients have at least two visits and their EHRs are complete. Here, we choose a real-world publicly available dataset MIMIC-III \cite{MIMIC} \footnote{https://mimic.physionet.org}, in which patients stayed within the intensive care units (ICU) at Beth Israel Deaconess Medical Center and had relatively complete health records with multilevel heterogeneous data. Though MIMIC-III belongs to the ICU data, there are certain patients with multiple visits. Hence, we utilize it as our experimental dataset.
Similar to \cite{Shang2019GAMENetGA}, we choose the medications prescribed by doctors for each patient within the first 24 hours as medicine set since it is usually a critical period for each patient to get rapid and accurate treatment \cite{Fonarow2005Effect}.
Besides, the medicine codes form NDC are transformed to ATC Level 3 for integrating with MIMIC-III.
Meanwhile, we employ the second hierarchy codes of the ICD9 codes\footnote{http://www.icd9data.com} as the disease category labels, since predicting category information not only guarantees the sufficient granularity of all the diagnoses but also improves the training speed and predictive performance \cite{Ma2017DipoleDP,Choi2017GRAMGA}.
For considering the laboratory results into decision-making process, we follow the feature extraction method used in \cite{Harutyunyan2019MultitaskLA}. Here, the time-window of each  laboratory indicator is 24 hours.
More information about the patients cohort from the dataset is listed in Table \ref{tab:dataset}.
\renewcommand\arraystretch{1.4}
\begin{table}
  \caption{Statistics of the MIMIC-III datasets}
  \vspace{0.2cm}
  \centering
  \begin{tabular}{l|c}
  \hline
  MIMIC III           & \multicolumn{1}{c}{Quantity} \\ \hline
  \# of patients                   & 4631 \\
  \# of unique diagnosis      & 1879 \\
  \# of unique medication     & 143 \\
  \# of unique laboratory indicators & 17 \\
  avg \# of visits            & 2.55 \\
  avg \# of diagnoses         & 10.16 \\
  avg \# of medications        & 7.33 \\\hline
  \end{tabular}
  \label{tab:dataset}
\end{table}

\subsection{Evaluation metrics}
\label{sec:metrics}
To evaluate the performance, we adopt the Jaccard Similarity Score (Jaccard), Precision Recall AUC (PR-AUC), Average Recall (Recall) and Average F1 (F1) as the evaluation metrics. Jaccard is defined as the size of the intersection divided by the size of the union of predicted set $\hat{Y}_{t}^{i}$ and ground truth set $Y_{t}^{i}$. Precision is used to measure the correctness of predicted medicines and Recall is used to measure the completeness of predicted medicines. F1 is often used as the comprehensive evaluation metric of prediction model.
\begin{equation}
\begin{split}
\text{Jaccard}=\frac{1}{\sum_{i}^{N} \sum_{t}^{T_{i}} 1} \sum_{i}^{N} \sum_{t}^{T_{i}} \frac{\left|Y_{t}^{i} \cap \hat{Y}_{t}^{i}\right|}{\left|Y_{t}^{i} \cup \hat{Y}_{t}^{i}\right|},
\end{split}
\end{equation}
where $N$ denotes the number of patients in test set and $T_{i}$ is the number of visits for the $i^{th}$ patient. Given
\begin{equation}
\begin{split}
\text{Recall}=\frac{1}{\sum_{i}^{N} \sum_{t}^{T_{i}} 1} \sum_{i}^{N} \sum_{t}^{T_{i}}
\frac{\left|Y_{t}^{i} \cap\hat{Y}_{t}^{i}\right|}
{\left|Y_{t}^{i}\right|}
\end{split},
\end{equation}
\begin{equation}
\begin{split}
\text{Precision}=\frac{1}{\sum_{i}^{N} \sum_{t}^{T_{i}} 1} \sum_{i}^{N} \sum_{t}^{T_{i}}
\frac{\left|Y_{t}^{i} \cap\hat{Y}_{t}^{i}\right|}
{\left|\hat{Y}_{t}^{i}\right|}
\end{split},
\end{equation}
the valuation metric F1 can be calculated as:
\begin{equation}
\begin{split}
\text{Jaccard}=\frac{1}{\sum_{i}^{N} \sum_{t}^{T_{i}} 1} \sum_{i}^{N} \sum_{t}^{T_{i}} \frac{2\times \text{Precision}\times \text{Recall}}{\text{Precision} + \text{Recall}}.
\end{split}
\end{equation}

\subsection{Benchmark methods}
To evaluate the effectiveness of the proposed model, it was compared to the following baseline methods:
\begin{itemize}[leftmargin=*]
    \item \textit{Nearest}. To predict treatment medicines for a patient $p_i$, Nearest was proposed to choose the treatment medications prescribed for patient $p_j$, who has the most similar historical laboratory indicators and medications with $p_i$.
    \item \textit{LR \cite{Luaces2012BinaryRE}}. It is a logistic regression with L1/L2 regularization. We sum the multi-hot vector of each visit together and apply the binary relevance technique \cite{Luaces2012BinaryRE} to handle multi-label output.
    \item \textit{NBN \cite{Alexiou2017ABM}}. Here, this method mainly utilizes the prior knowledge and employ the statistical methods to recommend corresponding medications for patients.
    \item \textit{Retain \cite{Choi2016RETAINAI}}. RETAIN is an interpretable model with a two-level reverse time attention mechanism to predict diagnoses, which can detect significant past visits and associated clinical variables. It can be used for similar sequential prediction tasks, such as predicting treatment medicines.
    \item \textit{DELSTM \cite{Jin2018KDD}}. This model utilizes additional input sequence as the input of a decomposed gate to control the memory cell state, which indirectly interacts with primary input sequence.
    \item \textit{PCLSTM \cite{Jin2018KDD}}. This structure takes all heterogeneous sequences as input. In other words, multiple sequences interact with each other via the neuron interactions in the way of concatenating both hidden states.
    \item \textit{RAHM \cite{AN2020103502}}. It builds a relation augmented hierarchical multi-task learning framework for learning multi-level relation aware patient representation for medication prediction.
    \item \textit{LEAP \cite{Zhang2017LEAPLT}}. Leap formulates the medicine prediction problem as a multi-instance multi-label learning problem, which mainly uses a recurrent neural network (RNN) to recommend medicines.
    \item \textit{DMNC \cite{Le2018DualMN}}. DMNC uses a memory augmented neural network to model the interaction of two asynchronous sequences for treatment prediction task \cite{Le2018DualMN}.
    \item \textit{GAMENet \cite{Shang2019GAMENetGA}}. It employs a dynamic memory network to save encoded historical medication information, and further utilizes a query representation formed by encoding sequential diagnosis and procedure codes to retrieve medications from the memory bank.
    \item \textit{AMANet \cite{He2020AttentionAM}}. AMANet leverages self-attention and inter-attention to capture the intra-view and inter-view interactions. Then it concatenates the information from history attention and dynamic external memory to predict the medications.
\end{itemize}
\subsection{Experimental settings}
\label{sec:settings}
We randomly split the patients in MIMIC-III dataset into training, validation and test sets with 2/3 : 1/6 : 1/6 ratios.
The random splitting and training processes were performed five times. Table \ref{tab:results_comparison} lists the results as averages across the five runs obtained for all the compared models in terms of the four evaluation metrics described in Section \ref{sec:metrics}
. Specificially, the embedding size and the hidden layer dimension for LSTM and GRU are all set as $128$ and $128$, respectively. The dropout rate is set as $0.4$, batch size is set as 10, and the mixture weights of objective function are set as $\eta = 0.99, \varepsilon = 0.01$. The values of attention sparse degree controlling parameters ${\gamma}_l$ in ASM are set as 1.5, 1.5, 1.3, respectively.
Training is done through Adam \cite{Kingma2014AdamAM} at learning rate 2e-4, and we report the model performance in test set within 40 epochs.
All methods are trained on an Ubuntu 16.04 with 64GB memory and Nvidia TAITAN XP GPU using the Pytorch 1.0 framework.

\subsection{Performance comparison}
\begin{table}
  \centering
  \caption{Performance Comparison of Benchmark Methods on MIMIC-III Dataset}
  \begin{tabular}{l|cccc}
  \hline
  Methods                              & Jaccard    & PR-AUC  & Recall   & F1 \\ \hline
  $\operatorname{Nearest}$          & 0.2019 & 0.2227 &0.3099 & 0.2899    \\
  $\operatorname{LR}$ \cite{Luaces2012BinaryRE}         & 0.3401	&0.5549	&0.4549 &0.4901 \\
  $\operatorname{NBN}$ \cite{Alexiou2017ABM}            & 0.3341	&0.5479	&0.5081 &0.4839 \\\hline
  $\operatorname{Retain}$ \cite{Choi2016RETAINAI}       & 0.3267	&0.5364	&0.4841 &0.4905 \\
  $\operatorname{DELSTM}$ \cite{Jin2018KDD}            & 0.3357	&0.5371	&0.5144 &0.5005 \\
  $\operatorname{PCLSTM}$ \cite{Jin2018KDD}            & 0.3301	&0.5109	&0.4952 &0.4940 \\
  $\operatorname{RAHM}$ \cite{AN2020103502}             & 0.3558	&0.5393	&0.5357 &0.5122 \\\hline
  $\operatorname{LEAP}$\cite{Zhang2017LEAPLT}          & 0.3111	&0.4212	&0.4541 &0.4627 \\
  $\operatorname{DMNC}$ \cite{Le2018DualMN}             & 0.3272	&0.4476	&0.5136 &0.5021 \\
  $\operatorname{GAMENet}$ \cite{Shang2019GAMENetGA}    & 0.3452	&0.4828	&0.5246 &0.5115 \\
  $\operatorname{AMANet}$ \cite{He2020AttentionAM}      &0.3641	&0.5595	&0.5547	&0.5381 \\\hline
  $\operatorname{MeSIN}_{DE}$ &0.3929&0.5633&0.5717&0.5624\\
  $\operatorname{MeSIN}_{Soft}$ &0.3941	&0.5663	&0.5798	&0.5636   \\
  $\operatorname{MeSIN}$      & \textbf{0.3975}	&\textbf{0.5684}	&\textbf{0.5934}	&\textbf{0.5670}  \\
  \hline
  \end{tabular}
  \label{tab:results_comparison}
\end{table}

As demonstrated in Table \ref{tab:results_comparison}, the benchmark models used in health informatics are divided into three categories: Shallow  methods, including Nearest, LR and NBN; Predictive models, including Retain, DELSTM, PCLSTM and RAHM; Recommendation models,including LEAP, DMNC, GAMENet and AMANet. From the table, we have an impressive observations that the proposed MeSIN achieves the superior performance over the listed benchmark models. Through detailed comparison with the benchmark models, several interesting observations can be made as follows.

First, as for the shallow methods, Nearest, LR and NBN achieved about at least 5.74\%, 1.35\%, 8.53\% and 7.69\% lower scores on medication recommendation task with respect to Jaccard, PR-AUC, Recall and F1 score, respectively, than MeSIN. On the one hand, this kind of method does not consider the temporality and heterogeneity of EHR data, and also overlook the relations between multiple sequences. On the other hand, Nearest achieves the worst performance which indicates that most of the patients possess distinct disease conditions within continuous admissions to hospital.

Second, as for predictive models in health informatics, Retain, DELSTM, PCLSTM and RAHM also achieve poor performance on medication recommendation task than MeSIN. We think that the main reason might be that they can not consider the current medical records including laboratory indicators and diagnosed diseases status, and only take the historical records into accounts. In addition, Retain is a two-level attention based model, that can capture temporal correlations and identify influential past visits. However, it can not consider all heterogeneous sequences respectively and can just concatenate the heterogeneous embeddings into one embedding which would confuse the embeddings obtained from different hierarchies of EHR data. DELSTM and PCLSTM mainly pay more attention on the sequences interaction in the temporal sequence learning process while overlook the inherent hierarchy structure of EHR data. In MeSIN, the multilevel learning framework is incorporated to extract useful information from such kind of inherent structure.
RAHM partially employs the hierarchy nature of EHR data to acquire better performance via multi-task learning framework.
However, it still employ single sequence learning methods to integrate with the historical information which might cause the confusion of different historical medical sequences.

Third, our proposed MeSIN outperforms all state-of-the-art methods used for medication recommendation such as LEAP, DMNC, GAMENet and AMANet about at most 8.64\%, 14.72\%, 13.93\% and 10.43\%, and at least 3.34\%, 0.92\%, 3.87\% and 2.89\% with respect to Jaccard, PR-AUC, Recall and F1 score. In practice, the medication recommendation problem within an admission might not be the pure sequential recommendation process and it also refers to the diverse correlations. Their poor performance could be attributable to the poor ability to capture such complicated correlations. Particularly, LEAP cannot capture the inherent multiple relations among heterogeneous sequences. While DMNC realizes the interactions of two sequences through attention based DNC blocks but neglecting the utilization of medications in the history visits. Similarly, AMANet also does not consider historical medications prescribed for patients, but it achieves relatively better performance through multiple attention networks for capturing the inter- and intra- correlations of heterogeneous sequences. However, AMANet neglects the captured evolution information such as disease progression through temporal sequence learning network, which is still a kind of important information in decision-making process.

Finally, we can observe the MeSIN also outperforms two special variants, $\operatorname{MeSIN}_{DE}$ and $\operatorname{MeSIN}_{Soft}$.
For the former variant, we replace the developed interactive LSTM network (InLSTM) in MeSIN with DELSTM network \cite{Jin2018KDD}.
In this variant, the interaction processes just consider the impact of auxiliary input on the memory cell state of primary input sequence learning network, and overlook the impact on the current input cell state. While the InLSTM network in MeSIN simultaneously considers the sequential interactions from above two aspects.
For the latter variant, we replace the incorporated attention weights computation method \textit{Entmax} in attentional selective module (ASM) of MeSIN with \textit{Softmax}.
In this variant, unlike \textit{Entmax}, \textit{Softmax} will generate the dense attention alignments that is wasteful, and can not pay more focus on the really important feature embeddings.

Therefore, the critical reasons that MeSIN achieves the best performance compared with all benchmark models can be summarized as follows:
(1) The multilevel learning framework can help capture the inherent causal relations of adjacent hierarchies;
(2) The incorporated multiple attentional selective modules in the framework realizes the effective embeddings selection and make the learned patient representation be more expressive;
(3) The designed InLSTM further reinforces the sequences interactions from both the historical memory cell and the input cell, which can further optimize the temporal sequence learning process by incorporating more useful calibrated information.

\subsection{Ablation study}
\label{ablation_study}

We now need to examine the effectiveness of different components in MeSIN and evaluate the contribution of different source data. Hence, we conduct two kinds of ablation studies respectively on model's components and multi-sourced EHR data.
\subsubsection{Model components}
\label{sec:ablation_model}

\begin{table*}[]
\centering
\caption{Performance Comparison of the variants of MeSIN on MIMIC-III Dataset}
\begin{tabular}{c|ccc|cc|c|cccc}
\hline
\multirow{2}{*}{Model} & \multicolumn{3}{c|}{ASM} & \multicolumn{2}{c|}{InLSTM} & GSFM & \multicolumn{4}{c}{Recommendation performance} \\ \cline{2-11}
 & \multicolumn{1}{c|}{Lab} & \multicolumn{1}{c|}{Diag} & Med & \multicolumn{1}{c|}{Diag} & \multicolumn{1}{c|}{Med} & Fusion & \multicolumn{1}{l|}{Jaccard} & \multicolumn{1}{l|}{PR-AUC} & \multicolumn{1}{l|}{Recall} & \multicolumn{1}{c}{F1} \\ \hline
$\operatorname{Vanilla}$ & \ding{55} & \ding{55} & \ding{55} & \ding{55} & \ding{55} & \ding{55} &0.3832 &0.5579&0.5451&0.5482  \\
$\operatorname{ASM}_{L}$ & \ding{51} & \ding{55} & \ding{55} & \ding{55} & \ding{55} & \ding{55} &0.3849&0.5592&0.5522&0.5501  \\
$\operatorname{ASM}_{LD}$ & \ding{51} & \ding{51} & \ding{55} & \ding{55} & \ding{55} & \ding{55} &0.3865&0.5595&0.5549&0.5546  \\
$\operatorname{ASM}_{LDM}$ & \ding{51} & \ding{51} & \ding{51} & \ding{55} & \ding{55} & \ding{55} & 0.3887 &0.5592&0.5684&0.5583  \\
$\operatorname{ASM}\_{\operatorname{InLSTM}}_{D}$ & \ding{51} & \ding{51} & \ding{51} & \ding{51} & \ding{55} & \ding{55} &0.3916&0.5643&0.5673&0.5619 \\
$\operatorname{ASM}\_{\operatorname{InLSTM}}_{DM}$ & \ding{51} & \ding{51} & \ding{51} & \ding{51} & \ding{51} & \ding{55}&0.3935&0.5651&0.5768&0.5639 \\\hline
$\operatorname{MeSIN}$ & \ding{51} & \ding{51} & \ding{51} & \ding{51} & \ding{51} & \ding{51} & \textbf{0.3975}	&\textbf{0.5684}	&\textbf{0.5934}	&\textbf{0.5670}  \\ \hline
\end{tabular}
\label{tab:ablation_comp}
\end{table*}

This ablation study is conducted to verify the effectiveness of different MeSIN components to its overall performance.
To determine whether the incorporated components improve the performance, we add them one by one from scratch and verify their performance by all evaluation metrics including Jaccard, PRAUC, Recall and F1 score.
Table \ref{tab:ablation_comp} presents the recommendation results of distinct MeSIN variants on the MIMIC-III dataset.
One of the basic baseline models, Vanilla, the medical codes are respectively added together as the enhanced embeddings in every module. Besides, the standard LSTM networks are also respectively employed as the temporal sequence learning networks in three distinct modules, and we employ the concatenation-based fusion method to replace the proposed global selective fusion module. However, Vanilla still achieves relatively better performance compared with benchmark models, which can attribute to the incorporation of multi-sources data and the integration of current laboratory results and diagnosed disease by concatenation-based fusion method.

\textbf{Attentional selective module (ASM)}.
As explained in Section \ref{sec:MeSIN}, ASM is introduced to automatically select the useful information and filter out noise information as much as possible by assigning corresponding attention weights to embeddings according to their respective importance. The following variants are tested to evaluate the contribution of ASMs from different modules to the overall performance of MeSIN:
\begin{itemize}[leftmargin=*]
    \item $\operatorname{ASM}_{L}$.
    In this variant, we incorporate an attentional selective module for laboratory results embeddings selection. The overall performance is slightly improved by 0.17\% on Jaccard in this case compared with the Vanilla model. This testifies that the introduced ASM module can help focus on the useful laboratory results embeddings by controlling the value of ${\gamma}_l$, by which we can obtain relatively better enhanced embedding as the input of temporal sequence learning network.
    \item $\operatorname{ASM}_{LD}$.
    Similar to $\operatorname{ASM}_{L}$, in this variant, we introduce an attentional selective module to replace the addition operation for diagnoses codes embeddings selection. In this way, the diagnoses codes embeddings that are irrelevant with the recommendation task would be discarded by sparse attention under the value control of ${\gamma}_d$. As a result, the performance of $\operatorname{ASM}_{LD}$ is improved by 0.16\% on Jaccard, which indicates the importance of the ASM in MeSIN in selecting the useful information from numerous medical codes embeddings.
    \item $\operatorname{ASM}_{LDM}$.
    In this variant, we further incorporate the third ASM module into the prescribed medications embedding module for selecting the most relevant historical medication codes embeddings to build the patient representation. As a result, $\operatorname{ASM}_{LDM}$ makes relatively better improvement compared with $\operatorname{ASM}_{L}$ and $\operatorname{ASM}_{LD}$. We think that it can attribute that the medication embedding module has direct relevance with the medication recommendation task. In the end, the incorporation of above three attentional selective modules bring about 0.55\% on Jaccard, 0.13\% on PR-AUC, 2.33\% on Recall, 1.01\% on F1 in total compared with Vanilla. But the important is that ASM makes MeSIN more interpretable by focusing on the really important features.
\end{itemize}

\textbf{Interactive Long-Short Term Memory network (InLSTM)}.
InLSTM is developed for reinforcing the interaction process of heterogeneous sequences,  which is beneficial to capture the correlations of sequences.
The following variants are tested to evaluate the contribution of InLSTM to the overall performance of MeSIN:
\begin{itemize}[leftmargin=*]
    \item $\operatorname{ASM}\_{\operatorname{InLSTM}}_{D}$.
    In this variant, we incorporate a novel InLSTM to replace the standard LSTM in $\operatorname{ASM}_{LDM}$ in diagnoses codes embedding module for enhancing the interaction process of disease progression and changing laboratory results. It achieves by 0.29\% on Jaccard compared with $\operatorname{ASM}_{LDM}$, which verifies the importance of considering the correlations of sequences such as between laboratory results and diagnosed diseases into the temporal sequence learning process.
    \item $\operatorname{ASM}\_{\operatorname{InLSTM}}_{DM}$.
    Here, the interactive LSTM is further introduced to the top hierarchy, prescribed medications embedding module for facilitating the medication codes sequential learning process. In this sequence learning network, the diagnosed diseases are utilized to enhance the interaction process with prescribed medications for providing complimentary useful information. Thus, the performance of $\operatorname{ASM}\_{\operatorname{InLSTM}}_{DM}$ outperforms the
   fifth variant $\operatorname{ASM}\_{\operatorname{InLSTM}}_{D}$ by 1.9\% on Jaccard, which further indicates the superiority of InLSTM in MeSIN than the standard LSTM in $\operatorname{ASM}_{LDM}$.
\end{itemize}

\textbf{Global selective fusion module (GSFM)}.
After step \textbf{\uppercase\expandafter{\romannumeral1}}, the hierarchically interactive temporal sequence learning procedure, the obtained multi-sourced embeddings are integrated together via proposed global selective fusion module (GSFM) for obtaining the patient representation. In this way, MeSIN can automatically learn the contribution scores of distinct embeddings to the medication recommendation task. As a result, it improves by 0.4\% on Jaccard compared with the sixth variant $\operatorname{ASM}\_{\operatorname{InLSTM}}_{DM}$. This also indicates the advantage of GSFM than the concatenation-based method used in above six variant models. However, owe to the utilization of concatenation-based fusion method, Vanilla gains relatively better performance than the benchmark methods.

\subsubsection{Heterogeneous Data}
\label{sec:ablation_data}
\begin{table}
  \centering
  \caption{Contribution of different data sources to MeSIN performance}
  \begin{tabular}{l|ccccc}
  \hline
  Methods  & Jaccard    & PR-AUC   &Recall  & F1      \\ \hline
  $\operatorname{NoLab}$ & 0.3861	&0.5612	&0.5526	&0.5567  \\
  $\operatorname{NoDiag}$ &0.3552	&0.5548	&0.5158	&0.5236 \\
  $\operatorname{NoMed} $ & 0.3859	&0.5587	&0.5514	&0.5537 \\
  $\operatorname{AllData} $ & \textbf{0.3975}	&\textbf{0.5684}	&\textbf{0.5934}	&\textbf{0.5670}   \\ \hline
  \end{tabular}
  \label{tab:ablation_data}
\end{table}

According to the proposed method, multilevel EHR data need to be input to MeSIN for obtaining the final patient representation. Though each of them plays a paramount role in the clinical decision-making scenario, here, we would build the following MeSIN variants to evaluate the impact of different heterogeneous data on medication recommendation results (Table \ref{tab:ablation_data}).
In $NoLab$, the laboratory results embedding module in MeSIN is removed, and the introduced $\operatorname{InLSTM}_d$ in diagnoses codes embedding module needs to be replaced by the standard LSTM network. In this case, patient's detailed health status is unknown.
In $\operatorname{NoDiag}$, the diagnoses codes embedding module is removed from MeSIN, and just retains the remain two modules. Under such circumstance, the learned patient representation will lose the key disease progression information.
In $\operatorname{NoMed}$, the medications codes embedding module is removed from MeSIN. In this way, the learned patient representation will lose the historical medications information.

Clearly, it can be noticed from Table \ref{tab:ablation_data} that the performance of variants all drop owing to some apparent reasons. First, in practice, most medications are prescribed conditioned on the diagnosed diseases. Therefore, the performance of $NoDiag$ drops dramatically, which validates the crucial role of diagnosed diseases and disease progression in medication recommendation task.
Second, though historical medications in ICU are not so much valuable for most patients, but are still a kind of important information in understanding patient's history diseases which can help know some detailed information such as allergic condition. Thus, the performance of $NoMed$ also drops significantly on medications recommendation task.
Third, the performance of $NoLab$ also drops significantly in this case but slightly compared with $NoDiag$. The main reason is that the current patient health status including diagnosed diseases and key laboratory indicators results is still a paramount indicator indicating patient's health status. Thus, though the importance of each hierarchy data within EHRs are diverse
from each other, all of them play important roles in medication recommendation task.
\subsection{Attention analysis in selective module}
\label{sec:att}

\begin{figure}
\centering
\includegraphics[width=1\linewidth]{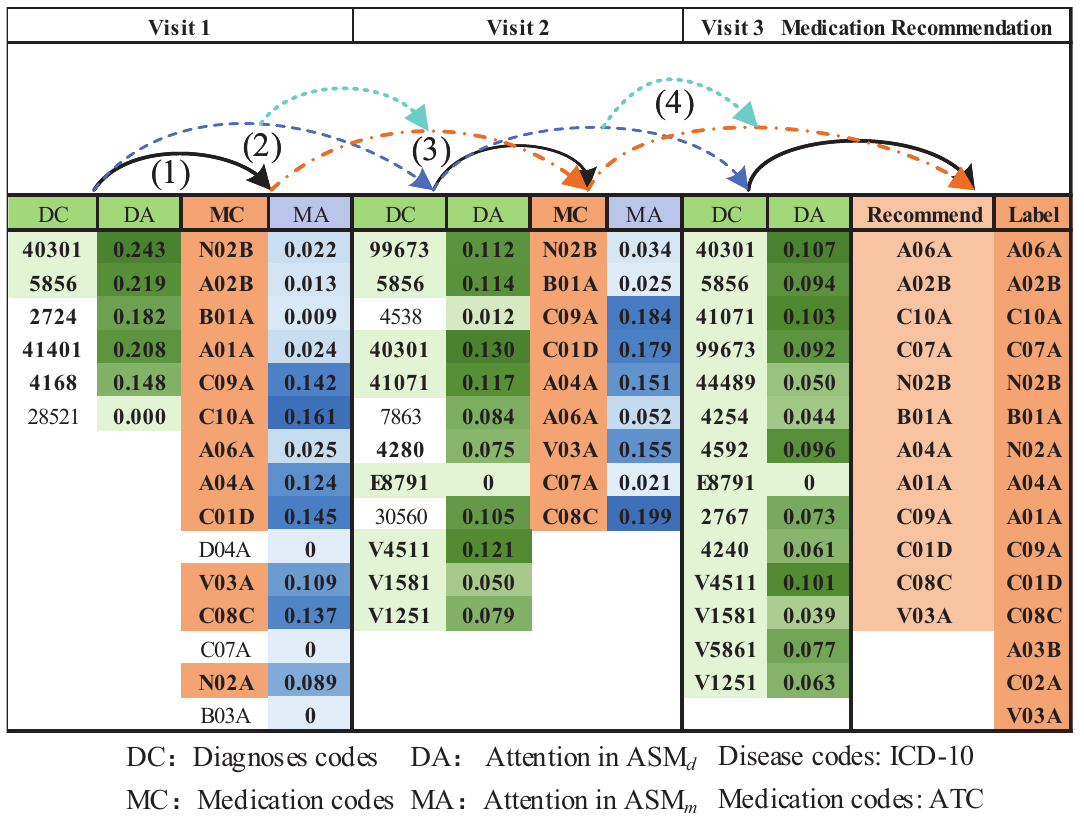}
\caption{The visualization of attentive process between the second and third hierarchy within our multilevel learning framework.}
\vspace{-0.2cm}
\label{Fig:AttAnalysis}
\end{figure}

\begin{figure*}
\centering
\includegraphics[width=0.8\linewidth]{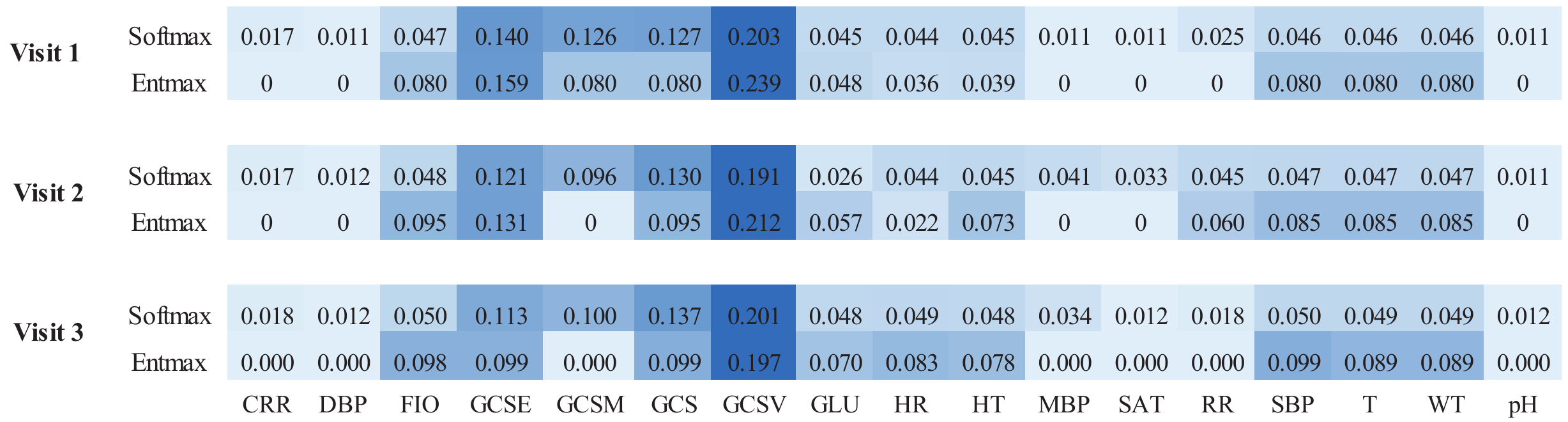}
\caption{The comparison of attention weights computed by Entmax and Softmax in laboratory results embeddings selection module.}
\vspace{-0.2cm}
\label{Fig:EntSoft}
\end{figure*}

\begin{figure}
\centering
\includegraphics[width=0.8\linewidth]{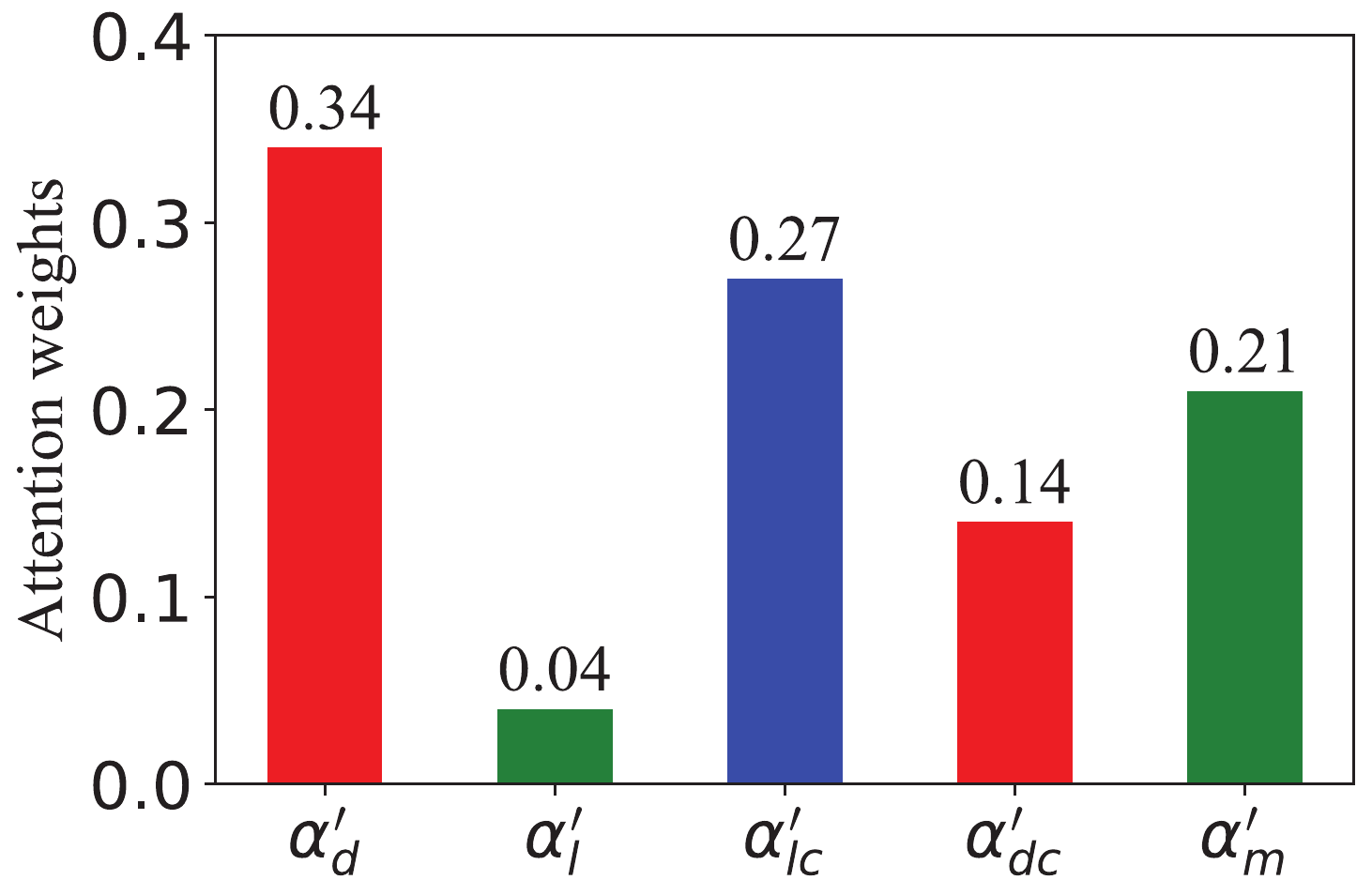}
\caption{Attention weights distribution in global selective module.}
\vspace{-0.2cm}
\label{Fig:GloablFusion}
\end{figure}

As discussed above, our newly developed MeSIN outperforms all benchmark models on medication recommendation for patients. Among the constituent components of MeSIN, the attentional selective module (ASM) plays a great role in the model, which has been testified through ablation studies about MeSIN in section \ref{sec:ablation_model}.
Actually, the positive influence of ASM should attribute to the selective ability of \textit{entmax}, which can increase focus on important medical codes embeddings and make the process more interpretable.
Hence, we perform attention analysis to explore the crucially attentive process shown in Figure \ref{Fig:AttAnalysis}, visualize the difference of softmax and entmax shown in Figure \ref{Fig:EntSoft}, and investigate the importance of multi-source embeddings shown in Figure \ref{Fig:GloablFusion}.

\textbf{The attentive process.} To clearly interpret the attentive process, as shown in Figure \ref{Fig:AttAnalysis}, we just consider the relations between the second and third hierarchies (diagnoses codes embedding module and prescribed medications embedding module) within our multilevel learning framework.
In addition, the quantitative value in column DA and column MA respectively denotes the attention weights calculated by Eq. (\ref{Eq:diag_ASM}) and Eq. (\ref{Eq:med_ASM}).
As shown in Figure \ref{Fig:AttAnalysis}, the attentive process can be categorized into four distinct but correlated processes. In this case, we have four interesting observations.
First, in the attentive process (1), the learned visit-level diagnoses codes embedding will be input into the medication codes embedding module for interacting with the medications codes embedding within historical visits. Thus, we can observe that there exists strong causal relations between \textit{DC} column (diagnoses codes) and \textit{MC} (medication codes) within each visit. In this way, the diagnoses codes that corresponds to the medications codes existing in the recommendation label column will be assigned more attention weights within each visit.
Second, owing to the temporal dependencies of EHR data, the recommended medications in \textit{Recommend} column not only depends on the diagnosed diseases in \textit{DC} column in the third visit, but also relies on the historical prescribed medications and diseases progression. As for this, as shown in the attentive process (2), the medication codes embeddings in historical visits but existing in \textit{label} column will be assigned more attention weights. Similarly, in the attentive process (3), as for the inherent causal relations between diseases and medications, the corresponding diagnoses codes embeddings will be also assigned more attention weights.
In the end, as shown in attentive process (4), for capturing the temporal dependency of EHR data, the medications codes sequence learning process will be influenced by the diagnoses codes sequence learning process under the help of proposed InLSTM in MeSIN.

\textbf{The calculation methods of attention weights: \textit{Softmax} and \textit{Entmax}.}
In MeSIN, \textit{Entmax} has been incorporated into the attentional selective module (ASM) to make more intelligent selections: assist to filter out noisy information and pay more focus on the important feature embeddings.
In figure \ref{Fig:EntSoft}, we provide a hot map which demonstrates the difference of attention weights computed by \textit{Entmax} \cite{Peters2019SparseSM} and \textit{Softmax} in laboratory results embeddings selection module.
As shown in figure, we observe that the calculation method \textit{Entmax} can generate sparse attention weights within each visit, in other words, it can make the attention scores of some unimportant indicator results embeddings to zero such as CRR, DBP, MBP, OS, RR and PH.
In this way, the MeSIN can intelligently make selections about which features embeddings are more important to be focused on and which are unnecessary to be payed so much attention on when making decisions in clinical decision-making process.
Therefore, such attention weights computation method in ASMs can help MeSIN increase focus on the really important features embeddings and make the model more interpretable.

\textbf{The importance of multi-source embeddings.}
As mentioned in section \ref{sec:GSF}, the global selective fusion module, to fuse the obtained five heterogeneous embeddings, we introduce a global selective fusion module, which can integrate them into patient representation according to respective importance score and minimize the adverse effect caused by noisy information.
In figure \ref{Fig:GloablFusion}, we can observe that ${\alpha}_{d}^{\prime}>{\alpha}_{lc}^{\prime}>{\alpha}_m^{\prime}>{\alpha}_{dc}^{\prime}>{\alpha}_{l}^{\prime}$ (see details in Eq.(\ref{Eq:global_att}-\ref{Eq:pat_fusion})), which indicates the importance ranking of multi-source embeddings. Such a phenomenon further testifies that diagnosed diseases especially the disease progression with historical disease information is the most important information for the medication recommendation task, which have been proved in the ablation study shown in Table \ref{tab:ablation_data}. The current laboratory result is the second critical factor when making decisions about the recommended medications. In addition, the historical prescribed medications are also taken into account. Finally, the historical laboratory results might be not so important in the intensive care unit (ICU). However, owing to that different patients might have different diseases status, the learned attention weights are also dynamically changing, which makes the computed relevance scores distribution are also diverse. For example, the historical medications might be more important than the diagnosed diseases when the diagnosis is adverse drug reaction.
Through the above analysis, we can see that MeSIN can provide some insightful and interpretable recommendation results.

\section{Conclusion}
In this paper, we propose a novel multilevel selective and interactive network for medication recommendation task with clinical EHR data.
In our model, the inherent causal relations and temporal dependencies of EHR data are effectively captured via proposed multilevel learning framework and a novel interactive LSTM cell. Considering the inevitable noise within EHR data, multiple attentional selective modules are incorporated into model for paying more focus on the really important feature embeddings and meanwhile provide insightful and interpretable recommendation results.
Finally, we evaluate our model on a real world and public clinical dataset. The experimental results show that our model achieves the best recommendation performance against eleven baselines in terms of Jaccard, PR-AUC, Recall and F1 score.
In the future, we plan to adapt the proposed approach for more healthcare prediction tasks based on sequential data and explore its usage in domains other than healthcare.

\section*{Acknowledgements}
Funding: This research was partially supported by the National Key R\&D Program of China (2018YFC0116800), National Natural Science Foundation of China (No. 61772110 and 71901011).

\section*{Declaration of Competing Interest}
Authors declare that there is no conflict of interest.


\bibliographystyle{cas-model2-names}

\bibliography{cas-refs}


\end{document}